\documentclass[lettersize,journal]{IEEEtran}
\usepackage{amsmath,amsfonts}
\usepackage{algorithmic}
\usepackage{algorithm}
\usepackage{array}
\usepackage[caption=false,font=normalsize,labelfont=sf,textfont=sf]{subfig}
\usepackage{textcomp}
\usepackage{stfloats}
\usepackage{url}
\usepackage{verbatim}
\usepackage{graphicx}
\usepackage{cite}
\usepackage{array}
\usepackage{tabularx}
\usepackage{multirow}
\usepackage{longtable}
\begin{document}

\title{Edge AI: Evaluation of Model Compression Techniques for Convolutional Neural Networks}

\author{
    Samer Francy,~\IEEEmembership{Member,~IEEE,} 
    Raghubir Singh,~\IEEEmembership{Member,~IEEE}
    \thanks{Samer Francy is with the Department of Computer Science, University of Bath, Bath, England (e-mail: samer.francy@bath.edu).}
    \thanks{Raghubir Singh is with the Department of Computer Science, University of Bath, Bath, England (e-mail: rs3022@bath.ac.uk).}
}

\maketitle


\begin{abstract}
This work evaluates the compression techniques on ConvNeXt models in image classification tasks using the CIFAR-10 dataset. Structured pruning, unstructured pruning, and dynamic quantization methods are evaluated to reduce model size and computational complexity while maintaining accuracy. The experiments, conducted on cloud-based platforms and edge device, assess the performance of these techniques. Results show significant reductions in model size, with up to 75\% reduction achieved using structured pruning techniques. Additionally, dynamic quantization achieves a reduction of up to 95\% in the number of parameters. Fine-tuned models exhibit improved compression performance, indicating the benefits of pre-training in conjunction with compression techniques. Unstructured pruning methods reveal trends in accuracy and compression, with limited reductions in computational complexity. The combination of OTOV3 pruning and dynamic quantization further enhances compression performance, resulting 89.7\% reduction in size, 95\% reduction with number of parameters and MACs, and 3.8\% increase with accuracy. The deployment of the final compressed model on edge device demonstrates high accuracy 92.5\% and low inference time 20 ms, validating the effectiveness of compression techniques for real-world edge computing applications.
\end{abstract}



\section{Introduction}
\IEEEPARstart{E}{dge} devices such as Internet of Things (IoT) are becoming increasingly important and widely used in our daily lives and industrial facilities. IoT is a network of things that empowered by sensors, identifiers, software intelligence, and internet connectivity, it can be considered as the intersection of the internet, things/objects (anything/everything), and data \cite{jabraeil_jamali_iot_2020}. The number of these devices is expected to increase even more \cite{voghoei_deep_2019}. These devices have the potential to perform complex Artificial Intelligence (AI) tasks locally, without relying heavily on cloud infrastructure \cite{nagaty_iot_2023}.
The rapid advancement of AI has led to the development of complex deep learning models that show high performance in different domains. Deploying AI models on edge devices has many advantages such as low latency, privacy and data security, bandwidth optimization, and reduced network dependence. Low latency is achieved due to real-time processing by instant data analysis on edge without waiting for remote server processing, this data analysis on the edge reduces transmitting data to the cloud which enhances security against breaches, reduces the bandwidth consumption, and reduces network dependence.

\subsection{Overview of Edge AI}

Edge AI represents a paradigm shift in the way AI is implemented in the context of the IoT. It capitalizes on the capabilities of IoT devices, enhancing real-time processing, analytics, and decision-making directly at the edge of the network. The IoT architecture, which is the foundation for Edge AI, typically involves three core layers \cite{jabraeil_jamali_iot_2020}. The layers are perceptual layer, where data is collected from various sensors and devices, network layer, where data is transmitted and routed through this layer, which is responsible for communication between devices and cloud services, and application layer, which processes and utilizes the data, providing insights and enabling actions.

\subsection{Convolutional Neural Networks (CNNs)}
CNN models are subsets of Deep Neural Networks (DNN) models. CNN models are effective for image and video-related tasks due to their ability to learn relevant features from the data by recognizing patterns, shapes, and structures in images, which is challenging for traditional machine learning models, that's why they are used for computer vision tasks such as image classification, object detection, and image segmentation \cite{bhatt_cnn_2021}. 

\subsubsection{CNN Architecture}
In general CNN models consist of below parts: 
\begin{itemize}
    \item \textbf{Input Image}:
    Pixels form the binary basis of computer images, while the human visual system operates through neurons with receptive fields. Similarly, CNNs function within their receptive areas, starting with simple patterns and advancing to more complex ones, making CNNs a promising tool for computer vision \cite{bhatt_cnn_2021}.
    \item \textbf{Convolutional Layer}:
    A convolutional layer in a CNN uses a small filter (e.g., 3x3 or 5x5) that slides over the input image. At each position, it multiplies its values with the overlapping image pixels and sums the results to produce an output. This sliding operation helps identify local features like edges and colors, building a hierarchical representation. The depth of the filter matches the input image's channels (e.g., 3 for RGB images). Stacking multiple filters allows the network to learn features at different abstraction levels \cite{bhatt_cnn_2021}.
    \item \textbf{Pooling Layer}:
    Pooling reduces the spatial size of feature maps. This not only lowers computational demands but also helps in extracting position and orientation-independent features essential for training. Two common pooling methods are maximum pooling and average pooling. In maximum pooling, a small kernel (e.g., 2x2) selects the maximum value within its range and places it in the output. In average pooling, a similar-sized kernel computes the average value within its area for each channel, maintaining the same depth. Pooling simplifies computation and weight requirements, with maximum pooling being the most commonly used method \cite{bhatt_cnn_2021}.
    \item \textbf{Activation Function}:
    The activation function, applies a mathematical operation to the filter's output to conclude the output of the network. The common choice is the Rectified Linear Unit (ReLU).  They fall into two categories, linear and non-linear activation functions \cite{bhatt_cnn_2021}.
    \item \textbf{Fully Connected Layer}:
    It functions as a feed-forward neural network (NN) typically situated at the network's lower layers. It receives input from the output of the last pooling or convolutional layer, which is flattened into a one-dimensional vector, enabling it to learn complex data relationships \cite{bhatt_cnn_2021}.
\end{itemize}

\subsubsection{Computation and Memory Demands}
\begin{figure*}
    \centering
    \includegraphics[width=\textwidth]{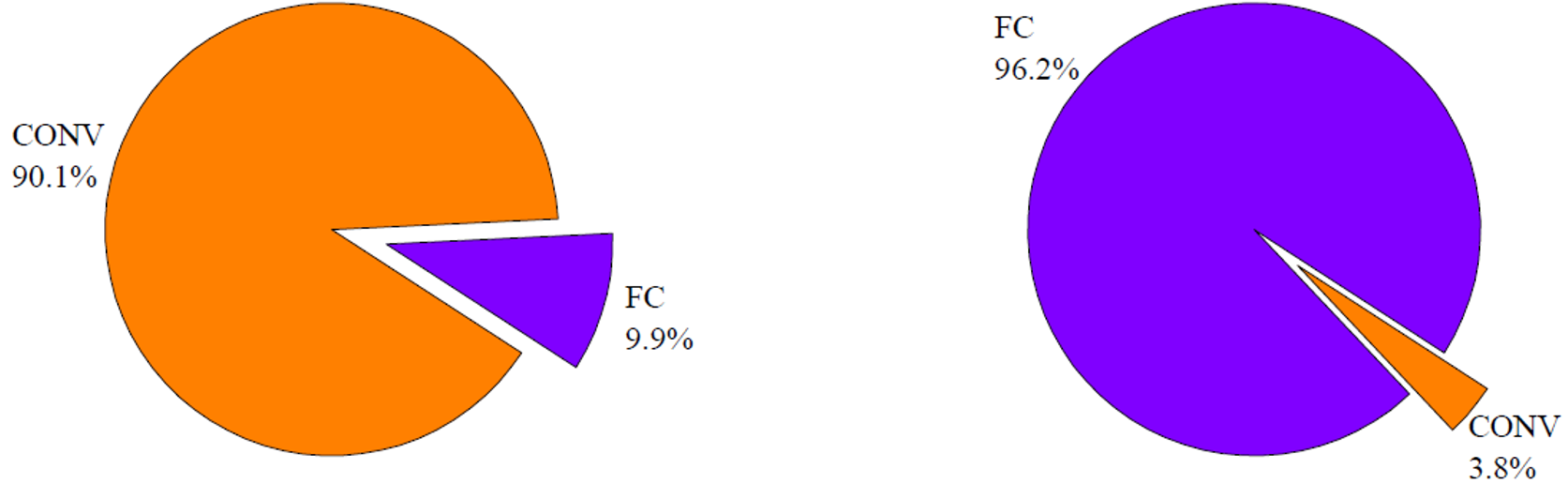} 
    \caption[Unbalanced Demand of CNN Layers]{Unbalanced Demand For Computation (Left) and Memory (Right) in AlexNet \cite{zhang_optimized_2019}.}
    \label{fig:1}
\end{figure*}

In CNNs, unbalance exists in resource demands between the layers. Convolutional layers primarily serve as feature extractors and heavily dominate the computational workload. In the case of AlexNet, for instance, the convolutional layers account for just 2 million weights but demand a substantial 1.33 Giga Operations Per Second (GOPS) of computation. In contrast, fully connected layers function as classifiers, accumulating information for high-level decisions, and bear the weight of the network with around 59 million parameters, yet they contribute significantly less to computation, requiring only 0.12 GOPS. This obvious contrast in resource allocation (Figure \ref{fig:1}) highlights the unbalanced demands between these two layers in CNNs \cite{zhang_optimized_2019}.

\subsubsection{Key CNN Architectures}
\begin{figure*}
    \centering
    \includegraphics[width=\textwidth]{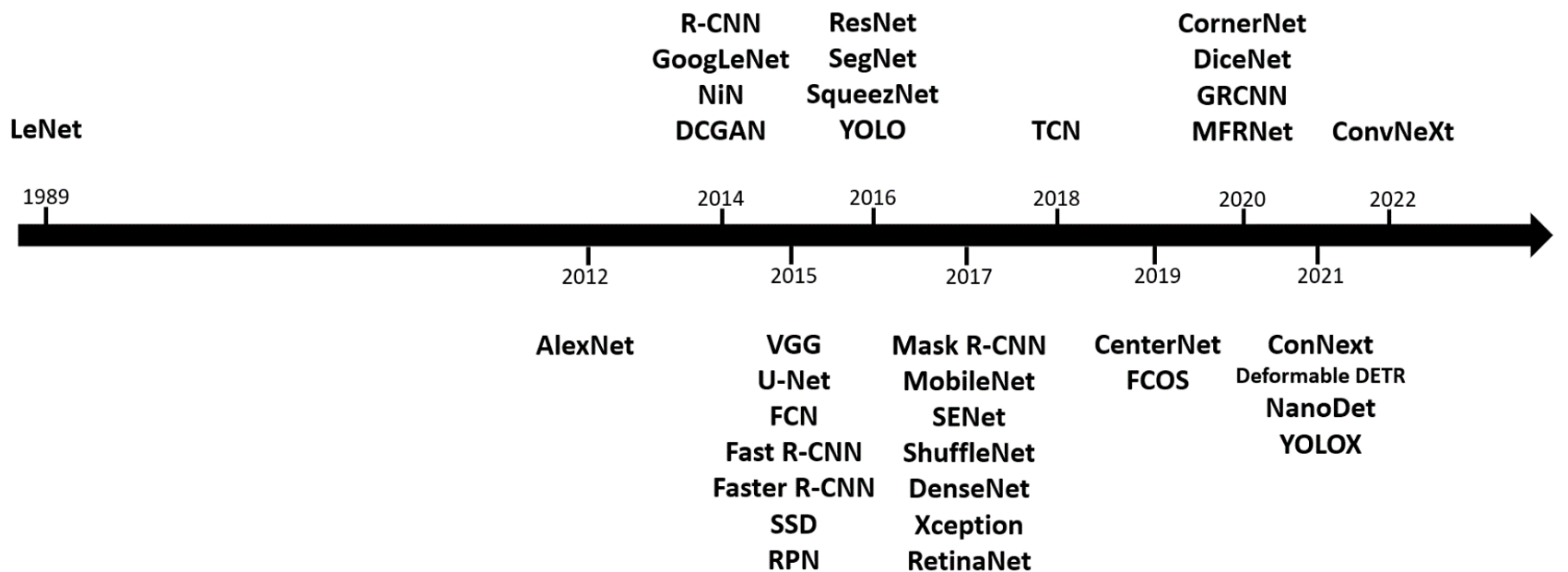} 
    \caption{Evolution of Key CNN Architectures Over Time.}
    \label{fig:2}
\end{figure*}

In 1989, the use of a NN architecture with convolutional layers for recognizing handwritten digits in the context of ZIP code recognition was introduced \cite{lecun_handwritten_1989}, That architecture consisted of input layer, 3 hidden layers, and output layer. Since then, CNN models have developed (Figure \ref{fig:2}) and became much deeper. 

\subsubsection{CNN on Edge}
Deploying CNN models on edge has a wide range of practical and industrial applications across various sectors. Here are some specific examples:
\begin{itemize}
    \item \textbf{Surveillance and Security}: It can perform real-time object detection and facial recognition for security monitoring, identifying intruders, and managing access control. Face identification was deployed using VGGFace \cite{meddad_hybrid_2020}. Video analysis was deployed using YOLOX \cite{joshi_object_nodate}. Infrared and Visible Image Fusion for security systems was deployed using DenseNet \cite{zhao_lightweight_2023}.  Human action recognition applications were deployed using ResNet \cite{amelio_representation_2023}.
    \item \textbf{Manufacturing and Quality Control}: It can inspect products on assembly lines for defects, ensuring quality control and minimizing errors. Real-time detection of steel strip surface defects was deployed using Faster R-CNN model \cite{ren_slighter_2018}. 
    \item \textbf{Agriculture}: Drones can monitor crops, detect pests, diseases, and nutrient deficiencies, enabling precision agriculture. Identifying rice leaf diseases in natural environments was deployed using GoogLeNet \cite{yang_googlenet_2023}. Pepper leaf disease identification was deployed using GoogLeNet \cite{dai_pepper_2023}. Detection for insect pests was deployed on YOLOX \cite{xu_asfl-yolox_2023}.
    \item \textbf{Healthcare and Wearables}: Wearable devices can continuously monitor vital signs, detect anomalies, and even diagnose certain health conditions. Medical diagnosis (Covid and Lung Disease Detection) was deployed using VGG, MobileNet, and AlexNet \cite{agarwal_efficient_2023}.  Automatically diagnose pneumonia and COVID-19 from chest X-ray images was deployed on DenseNet \cite{junior_optimization_2023}. Medical applications (e.g., COVID-19 detection, cardiomegaly diagnosis, brain tumor classification) were deployed using ResNet \cite{amelio_representation_2023}.
    \item \textbf{Energy Management}: It can monitor energy usage, optimize consumption patterns, and identify areas for energy efficiency improvements. Wind Turbine Maintenance and fault diagnosis was deployed using AlexNet \cite{gu_novel_2022}.
    \item \textbf{Environmental Monitoring}: It can monitor air quality, pollution levels, and weather conditions, providing valuable insights for urban planning. A smartphone app to perform fine-grained classification of animals in the wild was deployed using AlexNet, GoogLeNet, and ResNet \cite{tung_clip-q_2018}. Identification of mosquito species was deployed using AlexNet, DenseNet, Inception, ResNet, and VGG \cite{tang_position-based_2023}. 
    \item \textbf{Logistics and Inventory Management}: It can automate package sorting, inventory tracking, and warehouse management. Mobile robot to map its surroundings while detecting objects and people was deployed using AlexNet, GoogLeNet, and ResNet \cite{tung_clip-q_2018}. 
    \item \textbf{Autonomous Vehicles}: Vehicles can process real-time data from cameras and sensors using CNNs, aiding in autonomous navigation and collision avoidance. Instance objects detection system for intelligent service robots was deployed using Alexnet \cite{wang_object_2019}. Advanced driving assistance systems (ADASs) and automated vehicles (AVs) were deployed using Faster R-CNN \cite{chan_influence_2024}. 
\end{itemize}

Deploying CNNs on edge devices presents significant challenges mainly due to the limited computational resources, constrained memory, and power consumption constraints inherent to these devices. CNN models, known for their depth and complexity, often demand substantial computational power and memory, which may exceed the capabilities of edge hardware. Hence, compressing the model before deployment becomes imperative. Model compression techniques aim to reduce the size of the CNN model while preserving its performance, thereby enabling efficient utilization of computational resources and memory on edge devices. By compressing the model, we can mitigate the challenges associated with deploying CNNs on edge devices, ensuring that they can effectively perform tasks such as real-time image processing, object detection, and classification within resource-constrained environments.

With the enormous number of compression techniques proposed for CNNs, the rapid evolution of CNN architectures has created a \textbf{gap} in the field. This dynamic shift in architecture design requires an evaluation of existing compression methods, particularly in light of the demand to make these advanced CNN models suitable for deployment on edge devices. As CNN designs continue to advance, the challenge lies in adapting compression techniques to smoothly integrate with these modern architectures. This evaluation (either for each individual techniques or combined with each other) becomes important, as it not only ensures the continued relevance of compression techniques but also addresses the urgent need to make resource-intensive CNN models accessible and deployable on edge devices.

This work \textbf{aims} to evaluate CNN compression techniques that assure appropriate performance on edge devices. In the subsequent sections, this work reveals in a structured manner to evaluate the compression techniques for CNN models. section 2 provides a detailed review of related work, offering insights into prior research and establishing a foundational understanding of the topic. section 3 explains the methodology employed in conducting the experiments, describing the design and execution of the study. Following this, section 4 presents the experimental results and analyzes the findings to recognize trends and implications. Section 5 critically evaluates the results. Section 6 draws conclusions regarding the effectiveness and significance of the compression techniques. This organized framework aims to comprehensively explore and contribute to the field of model compression for efficient deployment in resource-constrained environments.


\section{Related Work}
Within the context of edge AI, it is important to address the critical need for model compression. The resource constrained nature of these devices requires more efficient AI models by minimizing memory and computational demands, ensuring faster inference speeds, and enhancing energy efficiency. Below will explore various model compression techniques and their implications for edge AI applications.

\subsection{Pruning}
Pruning is a key technique in DNN, aimed at enhancing efficiency and model generalization. It involves the removal of redundant components, such as parameters, neurons, filters, or entire layers, leading to several advantages. By reducing unnecessary parameters, it cuts down on storage requirements, and important for models deployed on devices with limited memory. Furthermore, it streamlines computational complexity during inference, resulting in faster predictions and lower power consumption. Pruning also mitigates overfitting by simplifying the network. Various pruning techniques, like weight pruning, neuron pruning, filter pruning, and layer pruning, offer different levels of granularity in component removal. Whether applied during or after training, pruning enables the creation of more compact and efficient CNN models tailored to specific needs, effectively balancing model size, computational efficiency, and accuracy. Weight pruning sets weight connections in the network to zero if they fall below a predefined threshold or are considered redundant. Neuron pruning focuses on removing entire neurons if they are found to be redundant. Layer pruning allows for the removal of entire layers that are considered less important \cite{choudhary_comprehensive_2020}.

\subsubsection{Pruning For Fully Connected Layer}
Fully connected layers are dense that makes the layer with high memory demand. Pruning them effectively reduces the memory burden and reduce size of the model.

\begin{figure*}
    \centering
    \includegraphics[width=\textwidth]{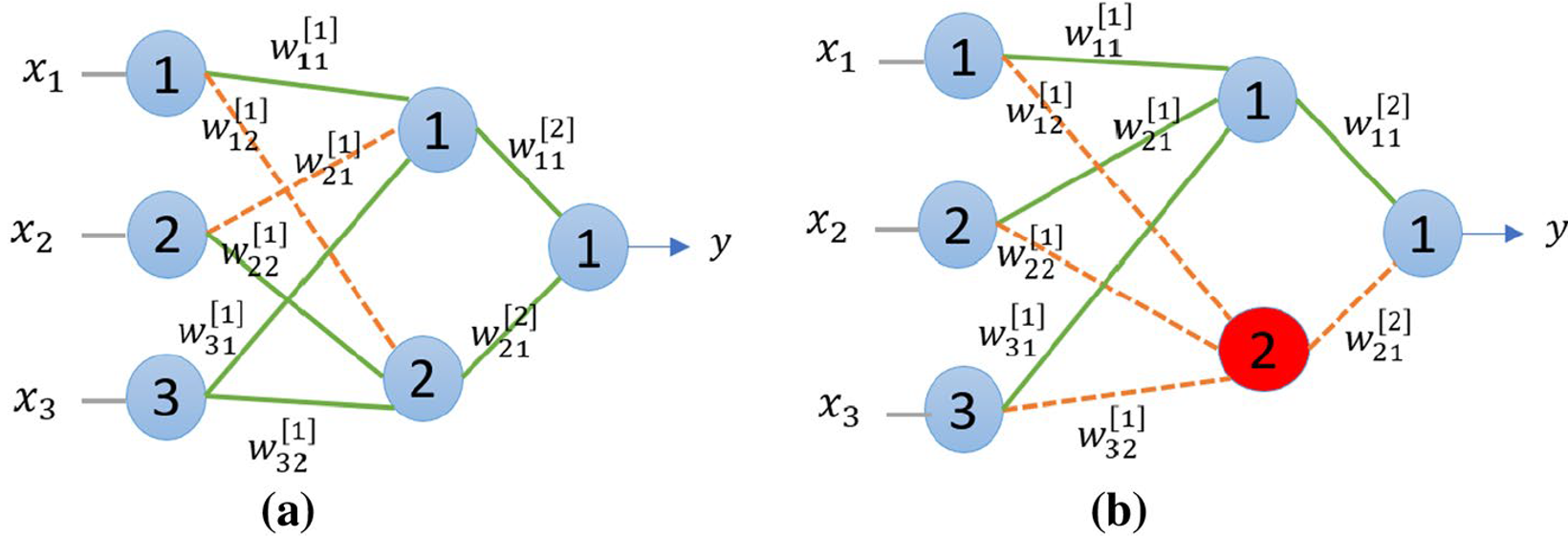} 
    \caption[Weight and Neuron Pruning]{Weight Pruning (a) and Neuron Pruning (b). x: input, w: weight. \cite{choudhary_comprehensive_2020}.}
    \label{fig:3}
\end{figure*}

It involves selectively removing weight connections and neurons to reduce the model's complexity while preserving performance. In a typical feed-forward NN, inputs are multiplied by corresponding weights, and a linear sum is computed at each neuron, which is then transformed by an activation function. As shown in Figure \ref{fig:3}, a network with 3 input neurons, 2 hidden neurons, and 1 output neuron may have multiple weight connections. Pruning can be applied to eliminate specific weight connections or entire neurons. By doing so, the total number of weight connections can be significantly reduced, leading to a more compact network. The concept of pruning was first introduced by \cite{lecun_handwritten_1989}, who proposed removing weights based on their saliency, with small-magnitude weights having less impact on training error. The process involves iterative retraining to regain accuracy, and the technique is known as 'Optimal Brain Damage (OBD)' where the second derivative of the objective function with respect to parameters is used to calculate the small saliency, facilitating informed pruning decisions. Since then, other pruning approaches have been introduced for fully connected layers \cite{choudhary_comprehensive_2020}.

\subsubsection{Pruning For Convolutional Layer}
Each convolutional layer typically consists of numerous filters that makes the layer with high computational demand. Pruning these less significant filters directly from the convolutional layer effectively reduces the computational burden and speeds up the model. Inspired by early pruning methods, new approaches have been introduced to be used to prune convolutional layers \cite{choudhary_comprehensive_2020}.

Bayesian was used to decide what to prune and the level of pruning, in this context involves employing scale mixtures of normals as priors for parameter weights in LeNet and VGG \cite{louizos_bayesian_2017}. Differential evolution based layer-wise weight method alongside three other pruning techniques (Naive Cut, Iterative Pruning, and Multi-Objective NN Pruning) was used to prune LeNet, AlexNet, and VGG16 \cite{wu_differential_2021}. Two fully connected layers are removed from the AlexNet architecture, and Batch Normalization (BN) is introduced to mitigate overfitting \cite{wang_object_2019}. Filters Similarity in Consecutive Layers (FSCL) for CNNs was used to reduce the number of filters while preserving important filters, ultimately improving model efficiency for VGG, GoogLeNet, and ResNet \cite{wang_filter_2023}. Structured pruning through sparsity-induced pruning was used to enhance the real-time implementation of the DEtection TRansformer (DETR) \cite{sun_pruning_2023}. Structured pruning was used to compress YOLOX, this process included sparse training to prune unimportant channels, with fine-tuning to recover accuracy \cite{xu_asfl-yolox_2023}. Evolutionary approach to filter pruning involved sequential application of multiple pruners in a specific order to sparsify LeNet and VGG-19 while maintaining model accuracy \cite{velayutham_c_evoprunerpool_2023}. Multilayer networks were used to represent and compress ResNets, it involved creating class networks, calculating arc weights, and forming a multilayer network. The overall degree of nodes in the multilayer network is used to select a subset of nodes for compression, and convolutional layers are pruned \cite{amelio_representation_2023}. To optimize the Fused-DenseNet-Tiny model for efficient detection of COVID-19 and pneumonia in chest X-ray images, three steps were implemented including removing insignificant weights, discarding pruning casings, and applying a compression algorithm \cite{junior_optimization_2023}. Deep Scalable Zerotree-based (DeepSZ) framework was used to address resource constraints by achieving significant compression for LeNet, AlexNet, and VGG while maintaining acceptable inference accuracy \cite{jin_deepsz_2019}. Compressing without retraining that was used with ResNet, AlexNet, VGGNet and SqueezeNet. It focused on convolutional and fully connected layers, while maintaining or improving classification accuracy \cite{dubey_coreset-based_2018}.

\subsection{Quantization}
Quantization plays an important role in addressing the resource-intensive nature of CNNs. By reducing the bit precision of model parameters, quantization not only conserves memory and energy but also enhances inference speed, making it an essential technique for deploying CNNs in resource-constrained environments such as edge devices. Weight clustering takes quantization to a more advanced level by organizing weights into clusters, where each cluster shares the same weight value. This approach minimizes the need for fine-tuning individual weights and can lead to substantial reductions in memory and computational overhead \cite{choudhary_comprehensive_2020}. 

Single Level Quantization (SLQ) and Multiple Level Quantization (MLQ) technique were used to quantize AlexNet, VGG, GoogleNet, and ResNet to the deployment of these models on resource-constrained mobile devices like mobile phones and drones \cite{xu_deep_2018}.

\subsection{Low-Rank Decomposition/Factorization}
It is a compression technique used with feed-forward NNs and CNNs, to reduce the size of weight matrices while preserving model performance. Singular Value Decomposition (SVD) is a popular factorization scheme that decomposes a weight matrix A into three smaller matrices: U, S, and $V^T$. U represents the left singular vectors, S is a diagonal matrix of singular values, and $V^T$ is the transpose of the right singular vectors. This factorization offers several advantages, such as reduced storage requirements, which is crucial for memory-constrained environments, and accelerated inference, especially in CNNs, as smaller matrices can be convolved faster. Low-rank factorization can be applied to fully connected and convolutional layers, making models more storage-efficient and faster without sacrificing performance. Careful selection of the rank is essential for achieving a balance between size reduction and model accuracy. Later, more approaches have been introduced \cite{choudhary_comprehensive_2020}.

Tucker decomposition for weight tensors was used to optimizes weight tensor dimensions of LeNet and ResNet models \cite{zhong_ada-tucker_2019}. Low-rank decomposition was used as an efficient method for compressing AlexNet, VGG, and ResNet without the need for fine-tuning to significantly reduce model size and computational complexity to make them more suitable for resource-constrained mobile and embedded devices \cite{zhang_speeding-up_2023}. Hardware-Aware Automatic Low-Rank Compression framework HALOC was used to compress ResNet, VGG and MobileNet, with the goal of efficiently exploring the structure-level redundancy in NNs by integrating principles from neural architecture search (NAS) \cite{xiao_haloc_2023}. Automatically Differentiable Tensor Network (ADTN) method was used to significantly reduce the number of parameters of fully connected NN, LeNet, and VGG while maintaining or enhancing the performance \cite{qing_compressing_2023}. Joint Matrix Decomposition, specifically Joint SVD (JSVD) was used to address the challenge of deploying ResNet with numerous parameters on resource-constrained platforms. It included Right JSVD, Left JSVD, and Binary JSVD algorithms \cite{chen_joint_2023}. Tensor Ring Networks (TR-Nets) was used as a method to effectively factorize LeNet and ResNet, thereby reducing computational and memory requirements \cite{wang_wide_2018}. Tucker decomposition with rank selection and fine tuning was used as a one-shot whole network compression scheme for deploying AlexNet, VGG, and GoogLeNet on mobile devices while maintaining reasonable accuracy \cite{kim_paraphrasing_2020}. Tensor Dynamic Low-Rank Training (TDLRT) was used to create a training algorithm with VGG and AlexNet that maintains high model performance while significantly reducing memory requirements for convolutional layers \cite{zangrando_rank-adaptive_2023}.

\subsection{Knowledge Distillation (KD)}
It is a technique used to transfer the knowledge learned by a larger, more complex model (the teacher model) to a smaller and lighter model (the student model). The primary goal of KD is to enable the student model to benefit from the generalization capabilities of the teacher model while being more lightweight in terms of parameters and computations. This technique helps to recover the accuracy drop occurs due to implementing other compression techniques. 

Knowledge transfer and distillation, initially introduced by \cite{bucilua_model_2006}, aimed to compress large ensemble models into smaller, faster counterparts with minimal performance loss. \cite{ba_deep_2014} extended this concept by empirically demonstrating that the intricate knowledge within larger DNNs could be effectively transferred to smaller, shallower models, yielding comparable accuracy. This involved training a large DNN and transferring its knowledge to a shallower network while minimizing the squared difference between the logits produced by the two models. These foundational ideas produced knowledge distillation, a widely used technique for training efficient models by transferring knowledge from larger ones. Later, more approaches have been introduced \cite{choudhary_comprehensive_2020}.

KD was used to improve the compression of LeNet and ResNet models when fresh training data is scarce, primarily through the use of synthetic data generated by Generative Adversarial Networks (GANs) \cite{liu_teacher-student_2020}. To fuse information from infrared and visible images while reducing DenseNet complexity and improving inference speed. Insights from pre-trained teacher models are transferred to the smaller student model \cite{zhao_lightweight_2023}. KD was used to develop a lightweight mosquito species identification model (EfficientNet) that balances efficiency and accuracy through the compression \cite{montalbo_automating_2023}.

\subsection{Mixed Techniques}
Different compression techniques are often combined and used together to achieve more effective and comprehensive model compression. Each compression technique targets specific aspects of the model, such as reducing model size, computation complexity, or memory footprint.

In-Parallel Pruning-Quantization CLIP-Q method combines network pruning and weight quantization was used to compress AlexNet, GoogLeNet, and ResNet \cite{tung_clip-q_2018}. Pruning and quantization were used to optimize the compression of AlexNet and reduce the number of parameters significantly while maintaining accuracy to be implemented on Field-Programmable Gate Array (FPGA) \cite{zhang_optimized_2019}. Pruning, quantization, and Huffman encoding combined with adversarial training were used to enhance the robustness and compression of AlexNet while also addressing the model vulnerability to adversarial attacks \cite{wijayanto_towards_2019}. Pruning and quantization were used to compress VGG and ResNet for remote sensing image classification, balancing computational complexity constraints while preserving model accuracy \cite{xie_co-compression_2023}. Low-rank decomposition and quantization were used to compress ResNet and MobileNet, and reduce the computational complexity while preserving high performance \cite{eo_effective_2023}. Pruning, quantization, and changing the model architecture were used to design a compact SqueezeNet with competitive accuracy while significantly reducing the number of parameters \cite{iandola_squeezenet_2016}. Quantization and pruning were used to develop an effective model compression framework for ResNet and MobileNet. The objective was to optimize the allocation of compression ratios to minimize performance degradation while reducing model size \cite{chen_towards_2023}. Joint quantization and pruning were used to develop a post-training model size compression method that efficiently combines lossy and lossless compression techniques to reduce the size of ResNet, MobileNet, RegNet, MNasNet, and YOLOv5 without sacrificing accuracy \cite{shi_lossy_2023}. 

\subsection{Other Techniques}
Depthwise separable convolutions was used to improve steel strip defect detection by creating a real-time and efficient model while maintaining high accuracy using Faster R-CNN \cite{ren_slighter_2018}. Deferential Evolution was used to develop an efficient and optimized AlexNet, VGG, and MobileNet for Covid and liver disease detection \cite{agarwal_efficient_2023}. Genetic Algorithm was used to reduce the storage space and inference time of VGG, ResNet, AlexNet, and SqueezeNet models \cite{agarwal_genetic_2023}. Factorization (changing kernel size) was used to improve the accuracy and computing efficiency of pepper leaf disease detection using GoogLeNet, specifically for the agricultural industry \cite{dai_pepper_2023}. Flexible and Separable Convolution (FSConv) was used to reduce computational costs without compromising the accuracy of VGG, ResNet, Faster R-CNN and RetinaNet \cite{zhu_fsconv_2023}. Efficient Layer Compression (ELC) was used to enhance the computational efficiency of VGG, ResNet, and ConvNeXt while preserving their representation capabilities \cite{wu_efficient_2023}.


\section{Design of The Experiments}
The experiments aimed to evaluate various compression techniques, namely pruning and quantization, on different types of ConvNext \cite{liu_convnet_2022} model. The experiments included training, fine-tuning, and evaluating of models using CIFAR-10 dataset. The setup involved conducting experiments both on cloud-based platforms and on edge devices to evaluate the performance of the compressed models.

\subsection[ConvNext]{ConvNeXt \cite{liu_convnet_2022}}
Is a modern CNN family produced as a journey of gradually modernize a standard ResNet toward the design of a vision Transformer. The journey starts from a ResNet-50 model, into a CNN architecture that mirrors some aspects of Transformers, particularly Swin Transformers. The roadmap:

\subsubsection{Training Techniques}
Vision Transformer training procedures were used to train ResNet-50 model, this included extending the training to 300 epochs (90 epochs originally), using AdamW optimizer, and data augmentation techniques (Mixup, Cutmix, RandAugment, Random Erasing, and regularization schemes including Stochastic Depth).

\subsubsection{Macro Design}
Number of blocks in each stage was adjusted from (3, 4, 6, 3) to (3, 3, 9, 3) and the stem was replaced with a patchify layer implemented using a 4x4, stride 4 convolutional layer (non-overlapping convolution).

\subsubsection{ResNeXt-ify}
 ResNeXt approach was adopted which is utilize grouped convolutions, where convolutional filters are divided into groups, each handling a subset of input channels, a variation of grouped convolution known as depthwise convolution was adopted, and the network's width was expanded by increasing the number of channels in the convolutional layers.

\subsubsection{Inverted Bottleneck}
\begin{figure}
    \centering
    \includegraphics[width=.5\textwidth]{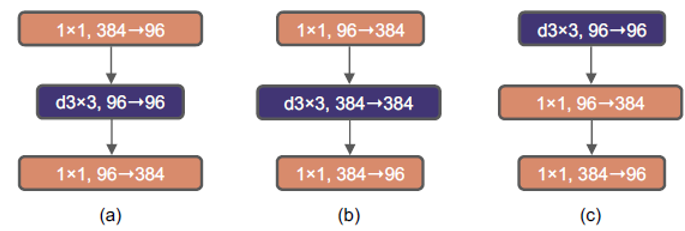} 
    \caption[ConvNeXt Block Modification]{Block modifications and resulted specifications. (a) is a ResNeXt block; in (b) we create an inverted bottleneck block and in (c) the position of the spatial depthwise conv layer is moved up \cite{liu_convnet_2022}.}
    \label{fig:4}
\end{figure}

The hidden dimension of the MLP block was changed to be four times wider than the input dimension as shown in Figure \ref{fig:4} (a and b)

\subsubsection{Large Kernel Sizes} 
The position of the convolutional layer is moved up and the kernel size was changed from (3x3) to (7x7) as shown in Figure \ref{fig:4} (a and c).
\subsection{Micro Design} 
\begin{figure}
    \centering
    \includegraphics[width=.4\textwidth]{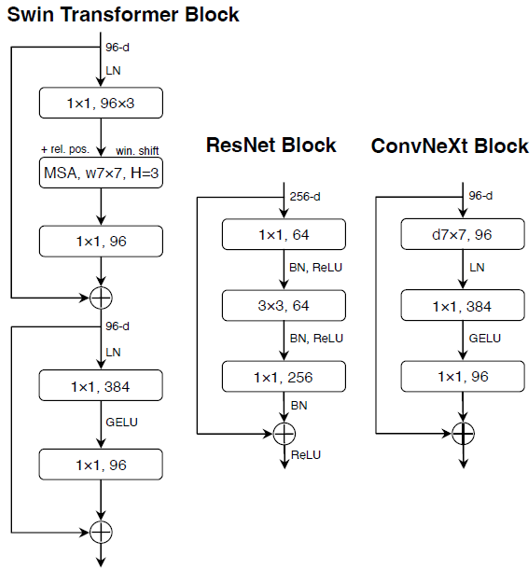} 
    \caption [Block Design] {Block designs for a ResNet, a Swin Transformer, and a ConvNeXt. Swin Transformer’s block is more sophisticated due to the presence of multiple specialized modules and two residual connections \cite{liu_convnet_2022}.}
    \label{fig:5}
\end{figure}

Replacing ReLU with Gaussian Error Linear Unit (GELU), fewer normalization layers, Substituting Batch Normalization (BN) with Layer Normalization (LN), and introducing separate downsampling layers as shown in Figure \ref{fig:5}. 

\subsection{Compression Techniques}
\subsubsection{Pruning}
Different pruning techniques have been used including structured and unstructured techniques.

\begin{itemize}
\item \textbf{Only Train Once (OTO) \cite{chen_otov3_2023}}:
OTO version 3 (OTOV3) is automated framework for structured pruning which involves removing entire structures or groups of parameters from a DNN. 
OTOv3 begins by analyzing the dependencies between the vertices of the target DNN. This analysis involves identifying accessory, Shape-Dependent (SD) joint, and unknown vertices that are adjacent and establishing their interdependencies. The goal is to form node groups based on these dependencies, laying the foundation for identifying interdependent vertices during structured pruning.

Using the information gathered from the dependency analysis, OTOv3 constructs a pruning dependency graph. This graph represents the interdependencies between vertices, with vertices in the same node group indicating their interdependency during structured pruning. The pruning dependency graph ensures the validity of the produced subnetwork by preserving essential connections between vertices.

OTOv3 partitions the trainable variables of the DNN into Pruning Zero-Invariant Groups (PZIGs) based on the pruning dependency graph. PZIGs consist of pairwise trainable variables grouped together, with each group representing a potential pruning structure. Node groups adjacent to the DNN output and containing unknown vertices are excluded from forming PZIGs to preserve output shapes and ensure model robustness as shown in Figure \ref{fig:6}.
\begin{figure*}
    \centering
    \includegraphics[width=\textwidth]{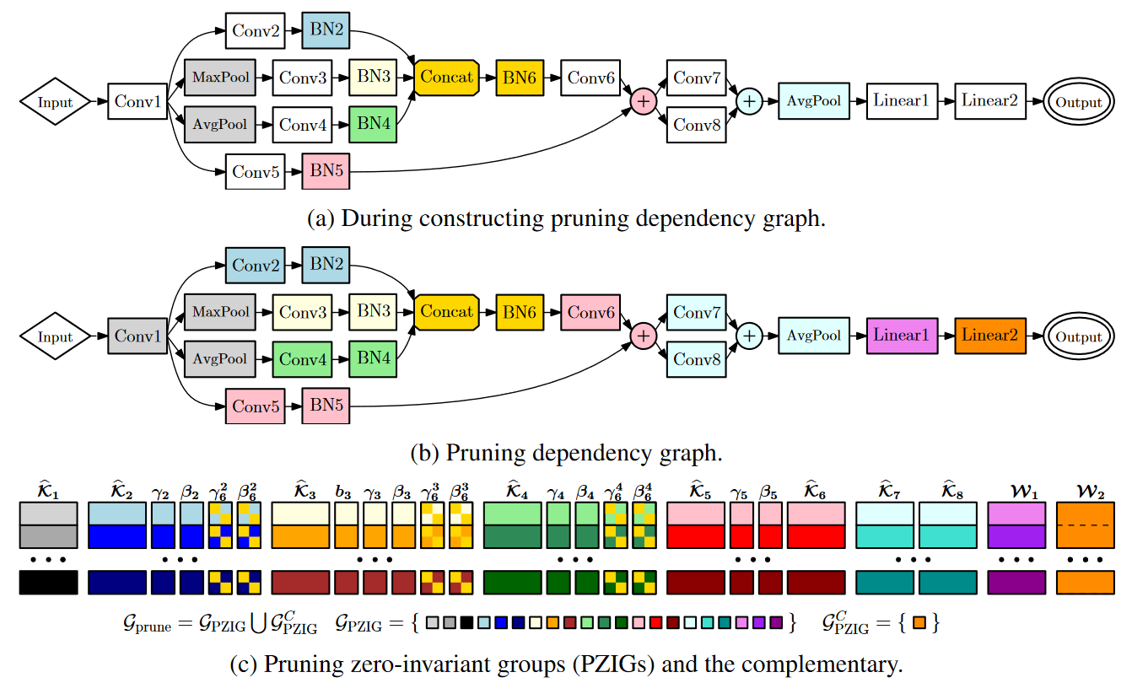} 
    \caption[Automated PZIG Partition]{Automated PZIG Partition \cite{chen_otov3_2023}.}
    \label{fig:6}
\end{figure*}

To jointly search for redundant pruning structures and train the remaining groups for optimal performance, OTOv3 employs the Dual Half-Space Projected Gradient (DHSPG) algorithm. DHSPG minimizes the objective function while introducing a sparsity constraint to identify redundant groups for removal. It employs saliency-driven redundant identification and a hybrid training paradigm to control sparsity and achieve better generalization performance as shown in Figure \ref{fig:7}.

\begin{figure}
    \centering
    \includegraphics[width=.45\textwidth]{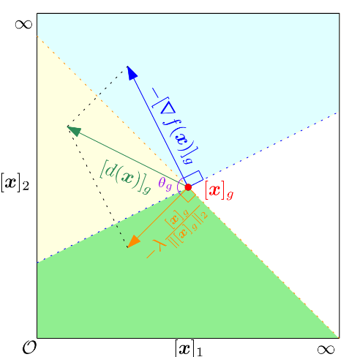} 
    \caption[Search Direction in DHSPG]{Search Direction in DHSPG \cite{chen_otov3_2023}.}
    \label{fig:7}
\end{figure}

\item \textbf{L1 Unstructured \cite{li_pruning_2017}}
L1 unstructured pruning is a technique used in machine learning, to reduce the size of neural networks by eliminating less important connections.
Each weight in the network is assigned a score based on its magnitude. This score reflects the importance of the weight in the network's performance. In l1 pruning, this score is often the absolute value of the weight.

A threshold is set, typically by selecting the top x\% of weights based on their magnitude scores. The threshold determines which weights will be pruned and which will be retained.

Weights that fall below the threshold are pruned, meaning they are set to zero and effectively removed from the network. This results in a sparser network architecture with fewer connections.

\item \textbf{Random Unstructured \cite{mittal_recovering_2018}}
Similar to l1 unstructured pruning, random unstructured pruning is also a technique used in machine learning, to reduce the size of neural networks by eliminating less important connections. The difference is the pruned weight are selected randomly instead of using l1 to decide the importance of the weights.

\end{itemize}

\subsubsection[Dynamic Quantization]{Dynamic Quantization \cite{xu_deep_2018}} 
Dynamic quantization is an approach aimed at optimizing the deployment of neural networks by reducing the precision of the weights. Unlike traditional quantization methods that apply a fixed quantization bit-width across all layers of the network, dynamic quantization adapts the quantization bit-width for each layer individually based on its representation abilities and capacities. This is achieved through the use of a bit-width controller module, which employs a policy gradient-based training approach to learn the optimal bit-width for each layer. By dynamically adjusting the quantization bit-width, dynamic quantization can strike a balance between maintaining accuracy and reducing memory size and computational costs.

\subsection[CIFAR-10]{CIFAR-10 \cite{noauthor_cifar-10_nodate}}
CIFAR-10 is a dataset used for computer vision and machine learning research, offering a rich resource for training and evaluating image classification algorithms. Comprising 60,000 32x32 RGB color images across 10 distinct classes (Airplane, Automobile, Bird, Cat, Deer, Dog, Frog, Horse, Ship, and Truck), CIFAR-10 facilitates comprehensive exploration of diverse visual concepts. With each class containing 6,000 images and a balanced distribution across the dataset, CIFAR-10 presents a well-structured foundation for model development. Its division into 50,000 training images and 10,000 test images, further segmented into multiple batches, enables strict evaluation and benchmarking of algorithms. In terms of computational requirements, CIFAR-10 generally requires less computation compared to CIFAR-100 and ImageNet due to its smaller image size and fewer classes which makes it suitable for experiments with limited computational resources.

\subsection{Experiment Setup}
Two types of experiments have been conducted, cloud-based experiments that focused on compressing the models and evaluating the techniques and edge-based experiment experiment to evaluate the performance of one of the compressed models.

\subsubsection{\textbf{Cloud-based Experiment Setup}}
\begin{figure*}
    \centering
    \includegraphics[width=\textwidth]{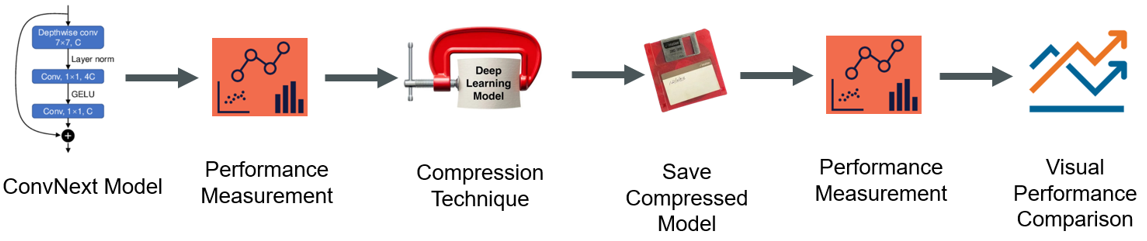} 
    \caption{Compression Evaluation Framework.}
    \label{fig:8}
\end{figure*}

Google Colab Pro+ was used to utilize GPU resources (NVIDIA A100 and V100 Tensor Core GPUs), facilitating accelerated model training and evaluation and background execution. The integration with Google Drive reduced overheads associated with uploading and downloading model data to and from cloud.
The evaluation framework was designed to profile the original model, compress it, profile the compressed model, and conduct comparison between the measurements before and after the compression as shown in Figure \ref{fig:8}.  

This profiling process involved measuring several key metrics:

\begin{itemize}

    \item \textbf{Accuracy}: The classification accuracy achieved by the model on the validation dataset.
    \item \textbf{Model Size}: The size of the model in megabytes (MB).
    \item \textbf{Number of Parameters}: The total count of trainable parameters in the model, measured in millions (M).
    \item \textbf{Number of MACs}: The number of multiply-accumulate operations performed during inference, measured in millions (M).
    \item \textbf{Number of Non-Zero Parameters}: The count of non-zero parameters in the model, essential for pruning-based techniques.
  
\end{itemize}

\subsubsection{\textbf{Edge-based Experiment Setup}}
A compressed model was deployed on edge with CPU (11th Gen Intel(R) Core(TM) i7-1165G7 @ 2.80GHz   2.80 GHz), RAM (16GB), and laptop integrated camera.

2 samples from each of CIFAR-10 classes have been selected randomly from the internet, printed on A4 papers, and placed in front of the camera to measure the accuracy and the inference time.


\section{Running The Experiments and Experimental Results}
\subsection{Cloud-Base Experiments}
different experiments have been conducted on cloud to evaluate different compressing techniques and different versions of ConvNeXt model.
\subsubsection{Evaluate OTOV3 on Untrained Torch ConvNext Tiny, Small, Base, and Large}
\begin{figure*}
    \centering
    \includegraphics[width=\textwidth]{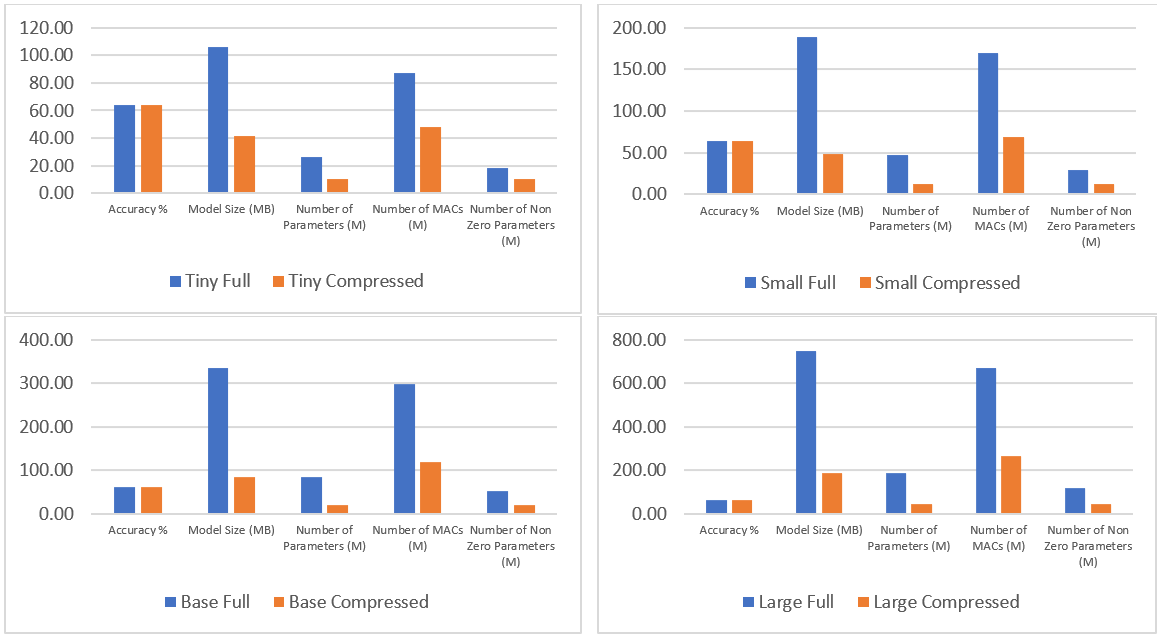} 
    \caption[OTOV3 Compression Performance]{OTOV3 Compression Performance with ConvNeXt Tiny, Small, Base and Large.}
    \label{fig:9}
\end{figure*}

Untrained ConvNeXt tiny, small, base, and large have been imported from Torch and been used to evaluate OTOV3 which train and prune at the same time, CIFAR-10 was used for training and evaluation, and 200 epochs were used for training and pruning. OTOV3 achieved high performance (Table \ref{table:1}) with reducing the model size (61\% for tiny and 75\% for small, base, and large), number of parameters (61\% for tiny and 75\% for small, base, and large), and MACs (45\% for tiny and 60\% for small, base, and large) as shown in Figure \ref{fig:9}. Meanwhile OTOV3 was able to increase both the full and compressed model accuracy through the training and pruning without any accuracy drop after pruning comparing to the full model. 

\begin{table*}[htbp]
    \centering
    \caption[OTOV3 Compression Numbers]{OTOV3 Compression Numbers with ConvNeXt Tiny, Small, Base and Large.}
    \label{table:1}
    \begin{tabular}{|>{\centering\arraybackslash}m{1cm}|>{\centering\arraybackslash}m{1.9cm}|>{\centering\arraybackslash}m{2cm}|>{\centering\arraybackslash}m{2cm}|>{\centering\arraybackslash}m{2cm}|>{\centering\arraybackslash}m{2cm}|>{\centering\arraybackslash}m{2cm}|}
    \hline
        \multicolumn{2}{|c|}{\textbf{Model}} & \textbf{Accuracy (\%)} & \textbf{Model Size (MB)} & \textbf{Number of Parameters (M)} & \textbf{Number of MACs (M)} & \textbf{Number of Non-Zero Parameters (M)} \\ \hline
        \multirow{2}{4em}{\textbf{Tiny}} &  Full &  63.81 &  106.26 &  26.53 &  86.88 &  18.41 \\
        \cline{2-7}
         & Compressed & 63.81 & 41.31 & 10.29 & 47.80 & 10.30 \\ \hline
        \multirow{2}{4em}{\textbf{Small}} & Full & 63.48 & 188.89 & 47.16 & 169.28 & 29.55 \\ 
        \cline{2-7}
         & Compressed & 63.48 & 48.04 & 11.94 & 68.24 & 11.96 \\ \hline
        \multirow{2}{4em}{\textbf{Base}} & Full & 61.22 & 334.28 & 83.50 & 299.20 & 52.22 \\ 
        \cline{2-7}
         & Compressed & 61.22 & 84.12 & 20.96 & 119.40 & 20.98 \\ \hline
        \multirow{2}{4em}{\textbf{Large}} & Full & 63.40 & 748.82 & 187.12 & 669.30 & 116.93 \\ 
        \cline{2-7}
        ~ & Compressed & 63.40 & 187.32 & 46.75 & 264.69 & 46.78 \\ \hline
    \end{tabular}

\end{table*}

\begin{figure*}
    \centering
    \includegraphics[width=\textwidth]{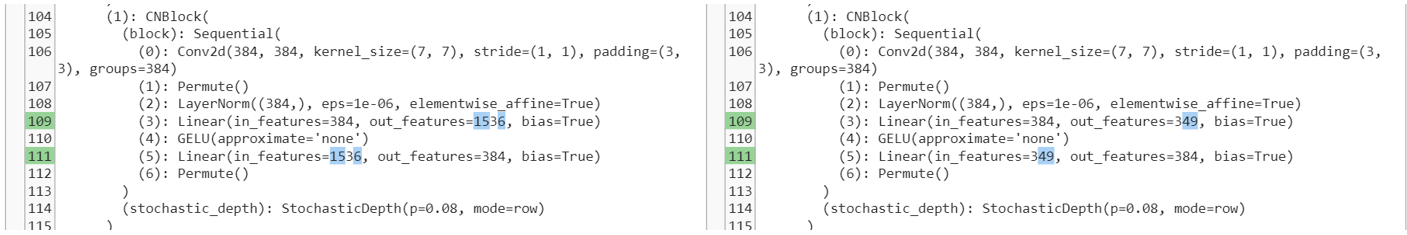} 
    \caption[OTOV3 Comparison Full vs Compressed]{Comparison Between ConvNeXt Small Full (Left) and Compressed (Right).}
    \label{fig:10}
\end{figure*}

To investigate the effect of OTOV3 on the model architecture, a comparison has been conducted between ConvNeXt small before and after compression. The Torch implementation of the model consist of many CNBlocks, each CNBlock consist of Conv2d, Permute, LayerNorm, Linear, GELU, Linear, and Permute layers. As shown in Figure \ref{fig:10}, OTOV3 reduced number of output features of the Linear layer (sequence 3) and the input features of the next Linear layer (sequence 5) and considering the big number of CNBlock in the model architecture, the reduction in model size and number of parameters after compression is justified as shown in Table \ref{table:2}.  

\begin{table}[htbp]
    \centering
    \caption[OTOV3 Comparison Full vs Compressed]{Comparison For Number of Output Features and Input Features of The Linear Layers in the CNBlocks Before and After OTOV3 Compression.}
    \label{table:2}
    \begin{tabular}{|>{\centering\arraybackslash}m{1.5cm}|>{\centering\arraybackslash}m{1.5cm}|>{\centering\arraybackslash}m{1.5cm}|>{\centering\arraybackslash}m{1.5cm}|}
    \hline
    \multicolumn{2}{|c|}{\textbf{Layer Index}} & \multicolumn{2}{|c|}{\textbf{ Input \& Output Features}} \\ \hline
        \textbf{Sequential} & \textbf{CNBlock} & \textbf{Full Model}  & \textbf{Compressed Model}  \\ \hline
        1 & 0 & 384 & 384  \\ \hline
        1 & 1 & 384 & 384  \\ \hline
        1 & 2 & 384 & 384  \\ \hline
        3 & 0 & 768 & 767  \\ \hline
        3 & 1 & 768 & 704  \\ \hline
        3 & 2 & 768 & 726  \\ \hline
        5 & 0 & 1536 & 251  \\ \hline
        5 & 1 & 1536 & 349  \\ \hline
        5 & 2 & 1536 & 242  \\ \hline
        5 & 3 & 1536 & 378  \\ \hline
        5 & 4 & 1536 & 293  \\ \hline
        5 & 5 & 1536 & 377  \\ \hline
        5 & 6 & 1536 & 340  \\ \hline
        5 & 7 & 1536 & 400  \\ \hline
        5 & 8 & 1536 & 394  \\ \hline
        5 & 9 & 1536 & 478  \\ \hline
        5 & 10 & 1536 & 414  \\ \hline
        5 & 11 & 1536 & 424  \\ \hline
        5 & 12 & 1536 & 410  \\ \hline
        5 & 13 & 1536 & 318  \\ \hline
        5 & 14 & 1536 & 488  \\ \hline
        5 & 15 & 1536 & 488  \\ \hline
        5 & 16 & 1536 & 402  \\ \hline
        5 & 17 & 1536 & 246  \\ \hline
        5 & 18 & 1536 & 402  \\ \hline
        5 & 19 & 1536 & 458  \\ \hline
        5 & 20 & 1536 & 323  \\ \hline
        5 & 21 & 1536 & 419  \\ \hline
        5 & 22 & 1536 & 446  \\ \hline
        5 & 23 & 1536 & 444  \\ \hline
        5 & 24 & 1536 & 441  \\ \hline
        5 & 25 & 1536 & 468  \\ \hline
        5 & 26 & 1536 & 1070  \\ \hline
        7 & 0 & 3072 & 208  \\ \hline
        7 & 1 & 3072 & 254  \\ \hline
        7 & 2 & 3072 & 262  \\ \hline
    \end{tabular}

\end{table}

\subsubsection{Evaluate OTOV3 on Untrained ConvNext Small (Torch vs. TIMM)}
\begin{figure}
    \centering
    \includegraphics[width=.5\textwidth]{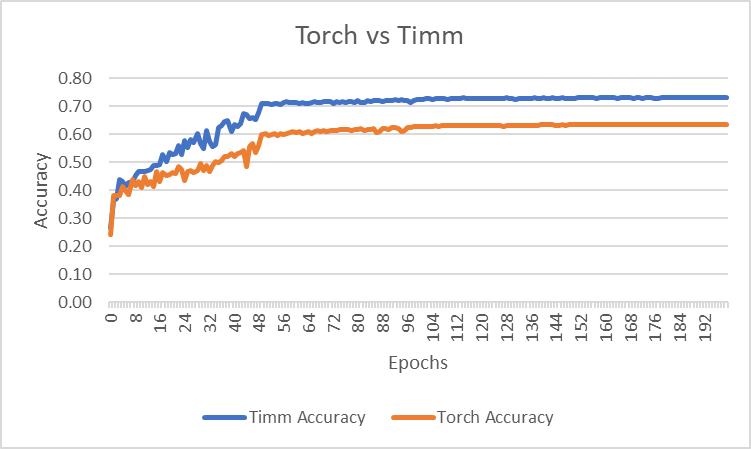} 
    \caption[OTOV3 Comparison Torch vs Timm]{OTOV3 Comparison Between Torch and Timm ConvNeXt Small.}
    \label{fig:11}
\end{figure}

Two untrained ConvNeXt small have been imported, one from Torch and one from TIMM \cite{wightman_rwightmantimm_2024} and been used to evaluate OTOV3 which train and prune at the same time, CIFAR-10 was used for training and evaluation, and 200 epochs were used for training and pruning. Although the compression performance was same with size reduction (75\%) but the accuracy after 200 epochs was less for Torch model (63\%) comparing to TIMM model (73\%) as shown in Figure \ref{fig:11}.

\begin{figure*}
    \centering
    \includegraphics[width=\textwidth]{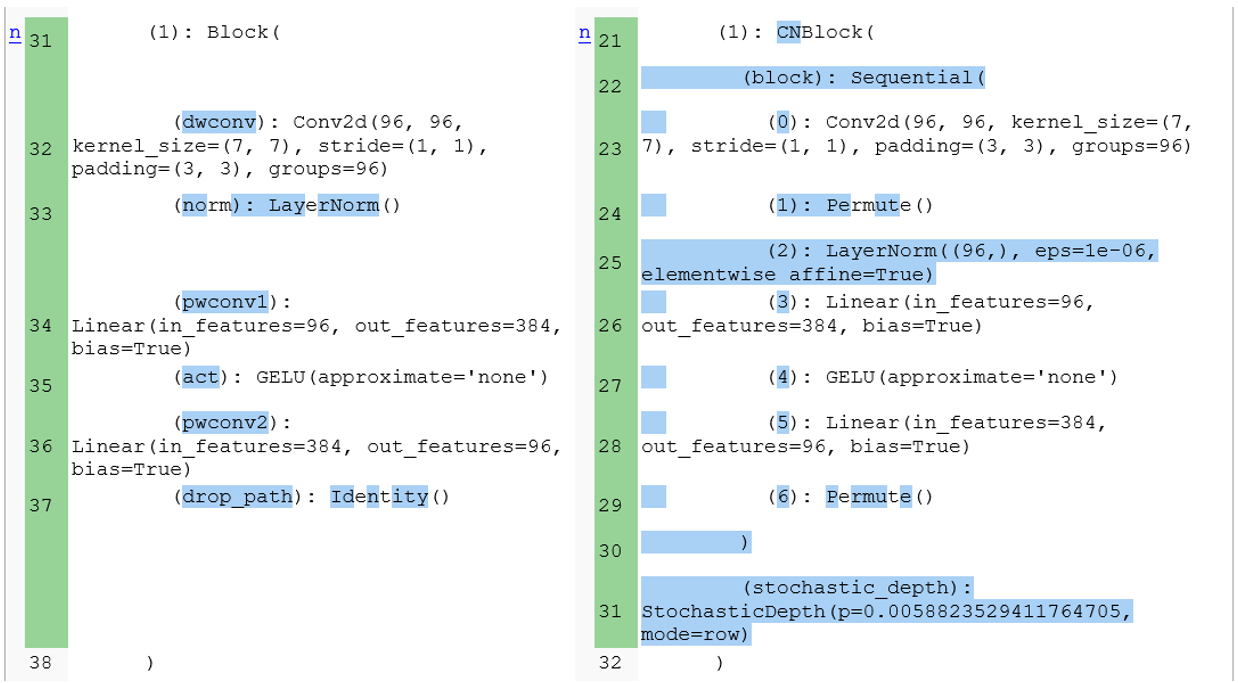} 
    \caption[Comparison Torch vs Timm]{Comparison Between ConvNeXt Small TIMM (Left) and Torch (Right).}
    \label{fig:12}
\end{figure*}

To investigate the accuracy performance of OTOV3 with Torch and Timm ConvNeXt Small, a comparison has been conducted between the two model architectures. The Torch model uses the CNBlock structure, which includes additional operations such as Permute and varying StochasticDepth probabilities. The TIMM model follows a simpler structure with Conv2d and LayerNorm, lacking the additional complexities introduced by CNBlock and associated operations in the Torch model as shown in Figure \ref{fig:12} which effects OTOV3 performance regarding the accuracy.

\subsubsection{Evaluate OTOV3 on Fine-Tuned Torch ConvNext Small}
\begin{figure}
    \centering
    \includegraphics[width=.5\textwidth]{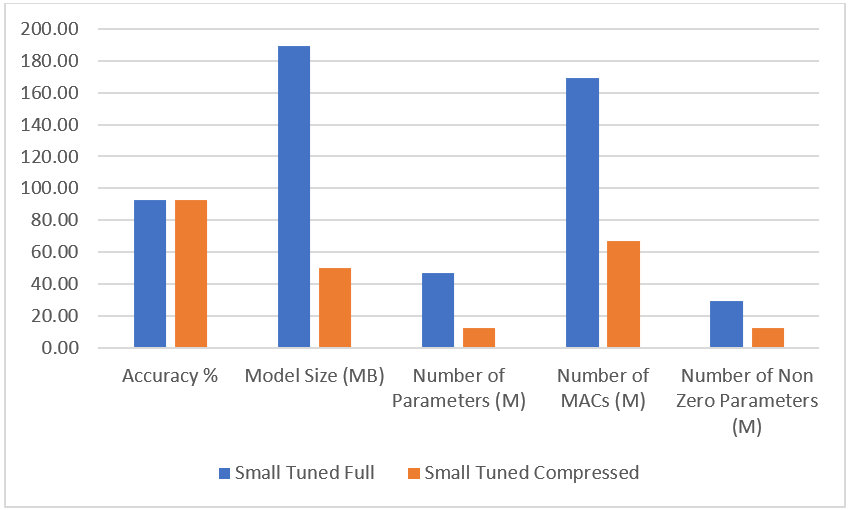} 
    \caption{OTOV3 Compression Performance with ConvNeXt Small Tuned.}
    \label{fig:13}
\end{figure}

A pre-trained ConvNeXt small have been imported from Torch and fine-tuned on CIFAR-10 with 100 epochs, the accuracy reached 89.5\%. This fine-tuned ConvNeXt small will be used for the rest of cloud-base experiments. This model was used to evaluate OTOV3, CIFAR-10 was used for training and evaluation, and 200 epochs were used for training and pruning. OTOV3 achieved high performance (Table \ref{table:3}) 74\% reduction in model size and number of parameters, 60\% reduction in MACs, and 3.8\% increase with accuracy as shown Figure \ref{fig:13}. The accuracy of the full model in (Table \ref{table:3}) (92.86\%) is different that the accuracy of the original model used in the experiment (89.5\%), that because OTOV3 trained the full model during the process which increased the model accuracy. 

\begin{table*}[htbp]
    \centering
    \caption{OTOV3 Compression Numbers with ConvNeXt Small Tuned.}
    \label{table:3}
    \begin{tabular}{|>{\centering\arraybackslash}m{1cm}|>{\centering\arraybackslash}m{1.9cm}|>{\centering\arraybackslash}m{2cm}|>{\centering\arraybackslash}m{2cm}|>{\centering\arraybackslash}m{2cm}|>{\centering\arraybackslash}m{2cm}|>{\centering\arraybackslash}m{2cm}|}
    \hline
        \multicolumn{2}{|c|}{\textbf{Model}} & \textbf{Accuracy (\%)} & \textbf{Model Size (MB)} & \textbf{Number of Parameters (M)} & \textbf{Number of MACs (M)} & \textbf{Number of Non-Zero Parameters (M)} \\ \hline
        \multirow{2}{4em}{\textbf{Small Tuned}} &  Full &  92.86 &	188.89 & 47.16 & 169.28 & 29.80 \\
        \cline{2-7}
         & Compressed & 92.86 & 50.03 & 12.44 & 67.41 & 12.46 \\ \hline
    
    \end{tabular}

\end{table*}

\subsubsection{Evaluate Unstructured Pruning}
\begin{figure*}
    \centering
    \includegraphics[width=\textwidth]{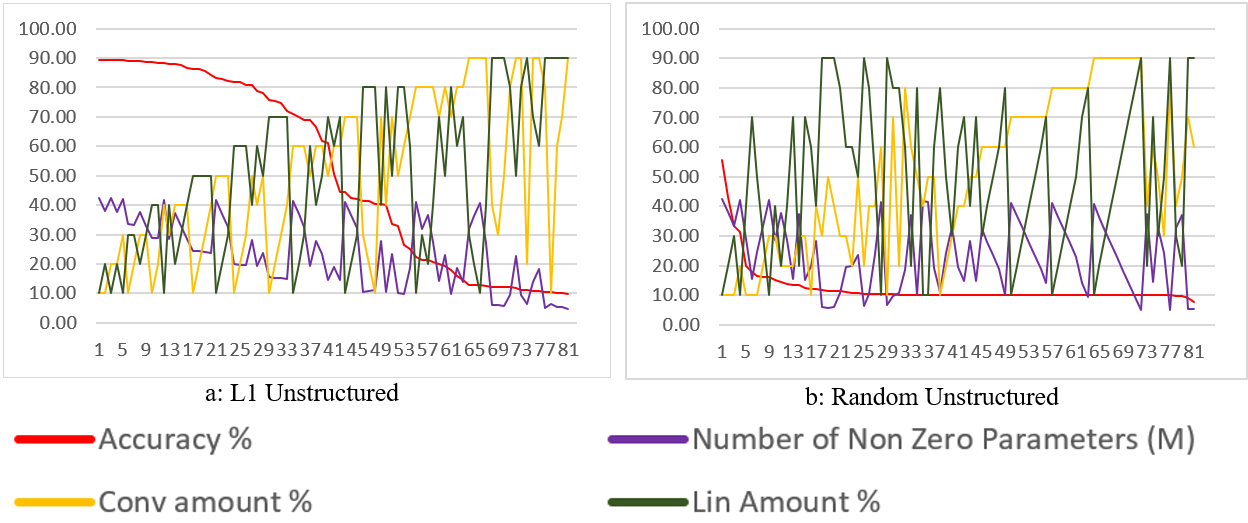} 
    \caption{Unstructured Pruning Performance with ConvMeXt Small.}
    \label{fig:14}
\end{figure*}

The Fine-tuned ConvNext Small was used to evaluate Pytorch L1 Unstructured Pruning and Random Unstructured Pruning by using different combinations of weights pruning percentages for linear (.1 to .9) and convolutional (.1 to .9) layers.  
In both experiments, the accuracy and the number of non-zero parameters were dropping as the values of weights pruning percentages for both linear and convolutional amounts were increasing as shown in Figure \ref{fig:14} a and b. Although the accuracy dropped but the model size, number of parameters, and MACs didn’t change as these techniques zero the weights instead of removing them.

\subsubsection{Evaluate Dynamic Quantization}
\begin{figure}
    \centering
    \includegraphics[width=.5\textwidth]{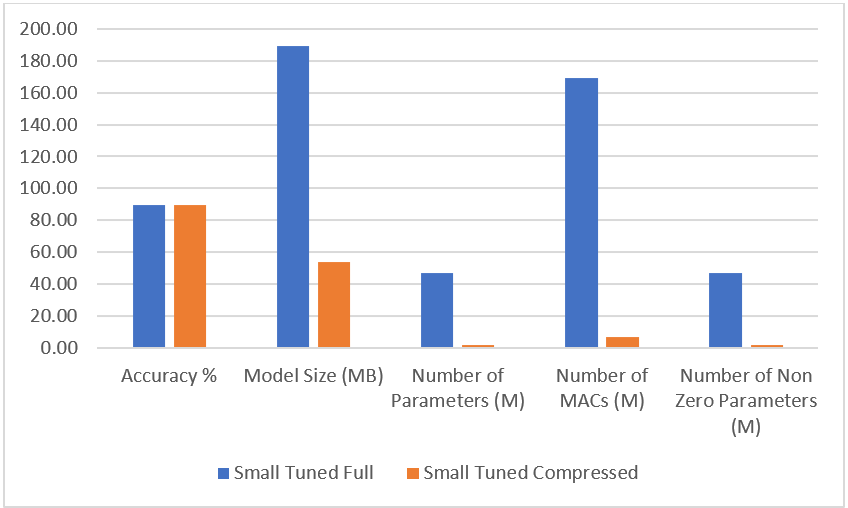} 
    \caption{Dynamic Quantization Compression Performance with ConvNeXt Small Tuned.}
    \label{fig:15}
\end{figure}

The Fine-tuned ConvNext Small was used to evaluate Pytorch dynamic quantization, 8-bit integer was used during the experiment and CIFAR-10 was used for evaluation.  Dynamic quantization achieved high performance (Table \ref{table:4}) 71\% reduction in model size, 95\% reduction with number of parameters and MACs, and 0.1\% drop with accuracy as shown in Figure \ref{fig:15}
\begin{table*}[htbp]
    \centering
    \caption{Dynamic Quantization Compression Numbers with ConvNeXt Small Tuned.}
    \label{table:4}
    \begin{tabular}{|>{\centering\arraybackslash}m{1cm}|>{\centering\arraybackslash}m{1.9cm}|>{\centering\arraybackslash}m{2cm}|>{\centering\arraybackslash}m{2cm}|>{\centering\arraybackslash}m{2cm}|>{\centering\arraybackslash}m{2cm}|>{\centering\arraybackslash}m{2cm}|}
    \hline
        \multicolumn{2}{|c|}{\textbf{Model}} & \textbf{Accuracy (\%)} & \textbf{Model Size (MB)} & \textbf{Number of Parameters (M)} & \textbf{Number of MACs (M)} & \textbf{Number of Non-Zero Parameters (M)} \\ \hline
        \multirow{2}{4em}{\textbf{Small Tuned}} &  Full &  89.53 & 188.89 & 47.16 & 169.28 & 47.17 \\
        \cline{2-7}
         & Compressed & 89.40 & 54.21 & 2.15 & 7.27 & 2.17 \\ \hline
    
    \end{tabular}
\end{table*}

\subsubsection{Evaluate Combination of OTOV3 and Dynamic Quantization}
Two compression stages were evaluated together as one compression pipeline, OTOV3 and Dynamic Quantization. The compressed ConvNeXt small model that was produced in experiment 4.1.3 (OTOV3) was furtherly compressed using Pytorch dynamic quantization using 8-bits integer (similar to experiment 4.1.5). The pruned models using OTOV2 \cite{chen_otov2_2023} had dependencies on Open Nural Network Exchange (ONNX) \cite{noauthor_onnx_nodate} which made it not applicable to be combined with other compression technique like quantization. In OTOv3, there was engineering changes produced the pruned model directly in Pytorch format, which enhanced the flexibility to be combined with quantization as this experiment did \cite{chen_otov3_2023}.

\begin{figure}
    \centering
    \includegraphics[width=.5\textwidth]{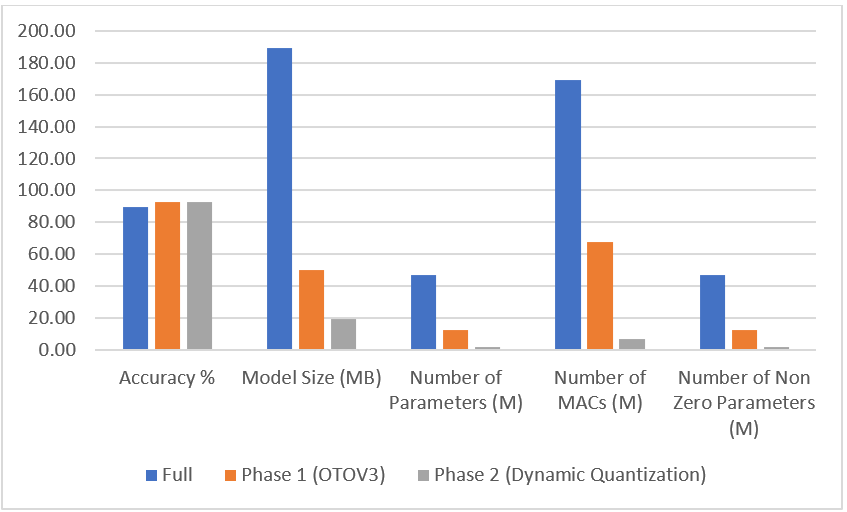} 
    \caption{OTOV3 and Dynamic Quantization Compression Performance with ConvNeXt Small Tuned.}
    \label{fig:16}
\end{figure}

Pruning using OTOV3 and Quantization using Pytorch dynamic quantization achieved high performance (Table \ref{table:5}) 89.7\% reduction in model size, 95\% reduction with number of parameters and MACs, and 3.8\% increase with accuracy as shown in \ref{fig:16}.

\begin{table*}[htbp]
    \centering
    \caption{OTOV3 and Dynamic Quantization Numbers with ConvNeXt Small Tuned.}
    \label{table:5}
    \begin{tabular}{|>{\centering\arraybackslash}m{4cm}|>{\centering\arraybackslash}m{2cm}|>{\centering\arraybackslash}m{2cm}|>{\centering\arraybackslash}m{2cm}|>{\centering\arraybackslash}m{2cm}|>{\centering\arraybackslash}m{2cm}|}
    \hline
        \textbf{Model} & \textbf{Accuracy \%} & \textbf{Model Size (MB)} & \textbf{Number of Parameters (M)} & \textbf{Number of MACs (M)} & \textbf{Number of Non Zero Parameters (M)} \\ \hline
          \textbf{Full} &  89.53 & 188.89 & 47.16 & 169.28 & 47.17 \\ \hline
          \textbf{Phase 1 (OTOV3)} & 92.86 & 50.03 & 12.44 & 67.41 & 12.46 \\ \hline
          \textbf{Phase 2 (Dynamic Quantization)} & 92.93 & 19.39 & 2.15 & 7.27 & 2.17 \\ \hline
    
    \end{tabular}
\end{table*}

\subsection{Edge-Base Experiment}
\begin{figure}
    \centering
    \includegraphics[width=.5\textwidth]{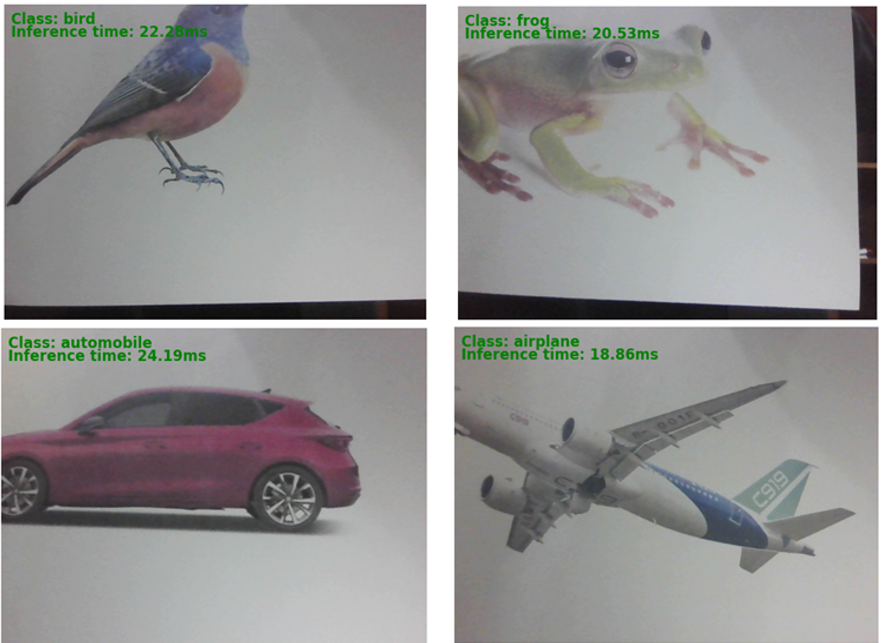} 
    \caption{Samples From the Compressed Model Output on Edge.}
    \label{fig:17}
\end{figure}

The final compressed ConvNeXt Small model in experiment IV.A.6 (OTV3 and Dynamic Quantization) was deployed on edge and the printed samples used to measure the accuracy and inference time by placing them in front of the camera. The compressed model achieved 92.5\% accuracy and 20ms inference time. Figure \ref{fig:17} shows samples of the output.  


\section{Analysis of Experimental Results}
Here, the experimental results obtained from running a series of experiments will be analyzed, the experiments aimed at evaluating various compression techniques applied to ConvNeXt models. The experiments were designed to evaluate the performance of techniques such as pruning and quantization on different sizes of ConvNeXt models, with a focus on both cloud-based compression experiments and edge-based deployment experiment. The analysis will be conducted with respect to the work's aim of evaluating CNN compression techniques that assure appropriate performance (size and inference time) on edge devices and resource-constrained environments.

\subsection{OTOV3 Evaluation on Untrained ConvNeXt Models}
The evaluation of OTOV3 on untrained ConvNeXt models demonstrated its effectiveness in achieving substantial compression while increasing model accuracy for both full and compressed models. Across varying sizes of ConvNeXt models, OTOV3 consistently produced impressive reductions in model size, number of parameters, and MACs, highlighting its ability to prune redundant structures effectively. This suggests that OTOV3 efficiently identifies and removes unnecessary parameters from the models, leading to more streamlined architectures without compromising predictive performance. However, an unexpected observation arose when comparing the accuracy of the Torch implementation with that of the TIMM implementation of ConvNeXt Small. The Torch implementation exhibited lower accuracy compared to its TIMM counterpart, indicating that OTOV3's performance regarding accuracy may be influenced by the details of the model architecture. This unexpected result suggests that different training strategies or adjustments may be necessary to optimize OTOV3's performance across various model implementations, emphasizing the importance of considering architectural differences when applying compression techniques like OTOV3.

\subsection{OTOV3 Evaluation on Fine-Tuned ConvNeXt Models}
When evaluating OTOV3's performance on a fine-tuned ConvNeXt model, notable improvements in compression performance were observed, confirming its effectiveness in reducing model size, parameters, and MACs while marginally enhancing accuracy. This outcome highlights the potential of integrating fine-tuning with structured pruning techniques to achieve even greater optimization of model performance. The fine-tuned model displayed enhanced compression capabilities compared to untrained models, suggesting that pre-training can significantly increase the effectiveness of compression techniques. This finding highlights the importance of leveraging pre-existing knowledge within models to maximize the benefits of compression, ultimately resulting in CNN models with higher performance.

\subsection{Unstructured Pruning Techniques}
During the evaluation of l1 unstructured and random unstructured pruning techniques, expected trends were observed regarding accuracy and compression. As the pruning percentages increased for both linear and convolutional layers, a corresponding decrease in accuracy was noted, while the model size, parameters, and MACs remained unaltered. This outcome aligns with the inherent nature of unstructured pruning, wherein weights are zeroed out but not entirely eliminated, resulting in sparse models without substantial reductions in computational complexity. However, the lack of significant reductions in computational complexity may constrain their effectiveness, particularly in resource-constrained environments where efficient utilization of computational resources is essential. This highlights the importance of considering the trade-offs between model compression and computational efficiency when selecting pruning techniques for deployment in real-world applications, especially in edge computing scenarios where computational resources are limited.

\subsection{Dynamic Quantization}
Dynamic quantization emerged as a highly effective technique for model compression, demonstrating remarkable reductions in model size, parameters, and MACs while preserving accuracy. Its effectiveness comes from the flexibility to adjust the quantization bit-width for each layer independently, leveraging the unique representation abilities and capacities of individual layers. This adaptability enables dynamic quantization to strike a balance between maintaining model accuracy and reducing memory size and computational costs effectively. Surprisingly, despite the substantial reduction in precision to 8-bit integers, the decrease in accuracy was minimal, suggesting the strength of dynamic quantization to quantization-induced errors. This unexpected resilience confirms that dynamic quantization is a reliable method for model compression, particularly in scenarios where memory and computational resources are limited, such as edge computing environments.

\subsection{Combination of OTOV3 and Dynamic Quantization}
The combination of OTOV3 with dynamic quantization represents an effective approach to CNN models compression, yielding impressive results in terms of both size reduction and accuracy improvement. By integrating structured pruning with quantization, this two-stage compression technique leverages the strengths of both methods to achieve synergistic effects in optimizing model efficiency. OTOV3, known for its ability to identify and remove redundant structures from neural networks while preserving accuracy performance, lays the foundation for subsequent compression stages. Dynamic quantization, on the other hand, dynamically adjusts the quantization bit-width for each layer, further reducing the memory footprint and computational demands of the model. The combined approach capitalizes on the flexibility of OTOV3 to seamlessly integrate with other compression techniques, enhancing its effectiveness in model optimization. The results of this experiment (Table \ref{table:6}) showcase the potential of multi-stage compression strategies in producing highly compressed models with improved accuracy, making them well-suited for deployment in resource-constrained environments such as edge devices.

\begin{table*}[!ht]
    \centering
    \caption{OTOV3 and Dynamic Quantization Reduction Numbers with ConvNeXt Small Tuned.}
    \label{table:6}
    \begin{tabular}{|>{\centering\arraybackslash}m{4cm}|>{\centering\arraybackslash}m{2cm}|>{\centering\arraybackslash}m{2cm}|>{\centering\arraybackslash}m{2cm}|>{\centering\arraybackslash}m{2cm}|>{\centering\arraybackslash}m{2cm}|}
    \hline
        \textbf{Technique} & \textbf{Accuracy Change} & \textbf{Size Reduction} & \textbf{Parameters Reduction} & \textbf{MACs Reduction} \\ \hline
        \textbf{OTOV3} & \textbf{3.75\%} & \textbf{73.51\%} & 73.62\% & 60.18\% \\ \hline
        \textbf{Dynamic Quantization} & -0.15\% & 71.30\% & \textbf{95.43\%} & \textbf{95.71\%} \\ \hline
       \textbf{OTOV3 and Dynamic Quantization} & \textbf{3.80\%} & \textbf{89.74\%} & \textbf{95.43\%} & \textbf{95.71\%} \\ \hline
    \end{tabular}

\end{table*}

\subsection{Edge-Based Experiment}
The edge-based experiment involved deploying the final compressed ConvNeXt model on edge devices, aiming to assess its performance in real-world deployment scenarios with limited computational resources. The results of the experiment were highly promising, demonstrating both high accuracy and low inference time for the compressed model. The high accuracy obtained indicates that the compression techniques employed, including OTOV3 and dynamic quantization, effectively preserved the model's accuracy performance despite reducing its size and computational complexity. This is crucial for ensuring the reliability and effectiveness of the model in practical applications where accurate predictions are essential. Furthermore, the low inference time achieved by the compressed model is equally significant, as it indicates efficient utilization of computational resources on edge devices, enabling rapid inference and response times. This is particularly important for latency-sensitive applications where quick decision-making is essential. Overall, the results of the edge-based experiment validate the suitability of the compressed ConvNeXt model for real-world deployment in resource-constrained environments, such as edge devices in IoT systems or mobile devices. 


\section{Conclusion and Future Work}
This work provides a detailed evaluation of various compression techniques for ConvNeXt models, concentrating on image classification tasks using the CIFAR-10 dataset. Through experiments conducted both in cloud environments and on edge devices, the research assessed the effectiveness of pruning and quantization methods in reducing model size and computational complexity while maintaining or improving accuracy.

Structured pruning, specifically OTOV3, demonstrated substantial reductions in model size, parameters, and MACs across different ConvNeXt model sizes, with some configurations even showing increased accuracy. The results also highlighted how differences in model architecture influence pruning efficiency, particularly in ConvNeXt Small variants. Extending the evaluation to fine-tuned models showed that combining pre-training with pruning further enhances compression efficiency, indicating significant potential for optimized model performance.

Unstructured pruning techniques, including L1 unstructured and random unstructured pruning, were examined but showed limited reductions in computational complexity, underscoring their limitations in resource-constrained environments. In contrast, dynamic quantization emerged as a highly effective compression technique, achieving significant reductions in model size and complexity without sacrificing accuracy. The combination of OTOV3 pruning with dynamic quantization demonstrated the potential for further enhancing compression performance, yielding models that are both compact and accurate.

The successful deployment of these compressed ConvNeXt models on edge devices validated their practical applicability, achieving high accuracy with low inference times. This confirms the relevance of the proposed techniques for real-world applications, particularly in environments where computational resources are limited.

Looking ahead, future research will explore additional pruning and quantization methods, such as Low-Rank Decomposition and Knowledge Distillation, to further mitigate accuracy drops post-compression. Additionally, extending this evaluation to larger datasets, including ImageNet and CIFAR-100, as well as exploring tasks beyond image classification, could provide deeper insights. These directions will contribute to advancing more efficient and effective model compression strategies, catering to the growing demands of edge computing and other resource-constrained scenarios.


\bibliographystyle{IEEEtran}
\bibliography{PaperBibFile}

\end{document}